\documentclass[a4paper,conference]{IEEEtran}
\IEEEoverridecommandlockouts
\ifCLASSINFOpdf
\else
\fi
\usepackage{textcomp} 
\usepackage{float}

\usepackage{graphicx, gensymb, url}
\usepackage{multicol, multirow}
\usepackage[dvipsnames]{xcolor}

\begin{document}
	
	\newcommand{\ArticleTitle}{3D attention mechanism for fine-grained classification of table tennis strokes using a Twin Spatio-Temporal Convolutional Neural Networks}
	
	\title{\ArticleTitle}
	
	
	
	%
	\author{\IEEEauthorblockN{Pierre-Etienne Martin\IEEEauthorrefmark{1},
			Jenny Benois-Pineau\IEEEauthorrefmark{1}
			\thanks{This work was supported by the New Aquitaine Region and Bordeaux Idex Initiative grant within the framework of the CRISP project - ComputeR vIsion for Sport Performance.},
			Renaud P\'{e}teri\IEEEauthorrefmark{2} and
			Julien Morlier\IEEEauthorrefmark{3}}
		\IEEEauthorblockA{\IEEEauthorrefmark{1}Univ. Bordeaux, CNRS,  Bordeaux INP, LaBRI, UMR 5800, F-33400, Talence, France\\
			Emails: pierre-etienne.martin@u-bordeaux.fr, jenny.benois-pineau@u-bordeaux.fr}
		\IEEEauthorblockA{\IEEEauthorrefmark{2}MIA, University of La Rochelle, La Rochelle, France\\
			Email: renaud.peteri@univ-lr.fr}
		\IEEEauthorblockA{\IEEEauthorrefmark{3}IMS, University of Bordeaux, Talence, France}}

	
	\maketitle
	
	\begin{abstract}
		The paper addresses the problem of recognition of actions in video with low inter-class variability such as Table Tennis strokes. Two stream, ``twin'' convolutional neural networks are used with 3D convolutions both on RGB data and optical flow. Actions are recognized by classification of temporal windows. We introduce 3D attention modules and examine their impact on classification efficiency. In the context of the study of sportsmen performances, a corpus of the particular actions of table tennis strokes is considered. The use of attention blocks in the network speeds up the training step and improves the classification scores up to $5$\% with our twin model. We visualize the impact on the obtained features and notice correlation between attention and player movements and position. Score comparison of state-of-the-art action classification method and proposed approach with attentional blocks is performed on the corpus. Proposed model with attention blocks outperforms previous model without them and our baseline.
	\end{abstract}

	\IEEEpeerreviewmaketitle
	
	\section{Introduction}
	\label{sec:intro}
	Action recognition in videos is one of the key problems in computer vision. Despite intensive research, recognition and discrimination of visually very similar actions remains a challenge~\cite{Dataset:AVA}. Several methods have been developed in the literature, the most recent ones using deep learning approaches~\cite{DBLP:journals/pami/RahmaniMS18}. Numerous works use models based on temporal networks such as RNN and LSTM~\cite{DBLP:journals/access/UllahAMSB18}. However, these networks are difficult to train and lack of stability~\cite{NN:Laptev18}. 3D convolutional neural networks are a good alternative for capturing long-term dependencies~\cite{NN:I3DCarreira}.
	They involve 3D convolutions: convolutions in space and time. These deep network architectures of multilayer perceptron type include - in the first layers - an extraction of features, and classifiers in the last layers. This type of approach translates in a powerful way through the prism of deep learning, what we knew beforehand: to extract features from time windows and use them for classification~\cite{Stoian16}.
	Recent methods also improve performances of Inflated 3D ConvNet~\cite{NN:I3DCarreira} by either capturing simultaneously slow and fast features using different video frame rates \cite{SlowFastNetworks} or by adding non local operations in the network \cite{NN:NonLocal}.	In \cite{NN:TSM}, the authors also obtain 3D CNN performances while keeping the 2D CNN complexity through Temporal Shift Modules.
	\par
	Recognition of similar actions is required quite frequently, and belongs to the fine-grained classification problem. In sport for instance, such as table tennis or gymnastics~\cite{Dataset:Gym},  exercises are filmed in the same environment and movements can be  quite similar. Hence the recognition problem becomes harder: the classifier cannot be helped by background information where an action is performed. The classifier has to focus on meaningful regions and changes in the video to be efficient. This is a subject of a recent trend in Deep Learning, that is the introduction of ``attention mechanisms''. The latter are designed to reinforce the contribution of meaningful features and channels into the decision and thus to increase the target accuracy.
	 Recently we proposed a comparative study of these attention mechanisms inherent to convolutional networks, as described in~\cite{Attention::saliency}. Selection of the most relevant characteristics in different layers is very similar to the human attention mechanisms measured in psycho-visual experiments. 
	While these attention mechanisms in 2D networks have been intensively studied~\cite{Attention::RANImageClassification}, this question remains to be further explored for a spatio-temporal content analysis using 3D convolutional networks. 
	\par
	The target application of our research is fine grained recognition of sport actions, in the context of the improvement of sport performances for amateur or professional athletes. Our case study is table tennis, and our goal is the temporal segmentation and classification of strokes performed. The low inter-class variability makes the task more difficult for this content than for more general action databases such as UCF-101~\cite{Dataset:UCF101} or DeepMind Kinetics~\cite{Dataset:Kinetics}. Twenty stroke classes and an additional rejection class have been established based on the rules of table tennis. This taxonomy has been discussed and designed with table tennis professionals. We are working on videos recorded at the Faculty of Sports of the University of Bordeaux (STAPS). The filmed athletes are students, and their teachers supervise the exercises performed during the recording sessions. These recordings are done without markers, which allows the players to play in natural conditions. The objective of table tennis stroke recognition is to help the teachers to focus on some of these strokes to help the students in their practice.
	\par
	We can mention that action recognition in Table Tennis videos is recently getting interest in the research community. In~\cite{TT:Tac-Simur}, the authors try to visualize and characterize tactics in table tennis competitions using a Markov chain model for comparing the profile of different players. Other works only focus on the ball tracking and trajectory estimation~\cite{TT:BallTracking&Trajectory}. In~\cite{TT:TTNet}, the authors propose an advanced real-time solution for scene segmentation, ball trajectory estimation and event detection but are not considering stroke classification.
	\par
	Table tennis strokes are most of the time visually similar. Action recognition in this case requires not only a tailored solution, but also a specific expertise to build the ground truth. This is the reason why annotations were carried out by professional athletes. They use a rather rich terminology that allows the fine-grained stroke definition. Moreover, the analysis of the annotations shows that, for the same video and the same stroke, professionals do not always agree. The same holds for defining temporal boundaries of a stroke, which may differ for each annotator. This variability cannot be considered as noise, but shows the ambiguity and complexity of the data that has to be taken into account. This new database, called \texttt{TTStroke-21} has been introduced in~\cite{PeCBMI,PeMTAP} as well as the twin network architecture used - Twin Spatio-Temporal CNN (TSTCNN).
	\par
	
	\par
     Attention mechanisms for action recognition have been recently introduced in LSTM~\cite{Attention::skeleton} in a recognition approach based on the analysis of joints of a human skeleton. In 3D CNNs both global channel attention and spatial attention maps for different feature layers have been proposed~\cite{Attention::3DRANs}. We also follow this trend and design attentional blocks for our TSTCNN model.
     In this paper, we propose spatio-temporal attention mechanisms in 3D convolution networks for recognition of challenging similar actions: Table Tennis strokes.
	\par
	The rest of the paper is organized as follows: in section~\ref{section:related_works}, works using attention mechanisms are presented. The section~\ref{section:method} presents the proposed method with attentional mechanisms and in details the attention block. The results are drawn in section~\ref{section:Results} through feature analysis and classification performances. The conclusion and perspectives are given in the section~\ref{section:conclusion}.
	
	\section{State of the Art on Attention mechanisms}
	\label{section:related_works}
	
	In this section, we present a brief state of the art on attention mechanisms introduced in convolutional networks for the classification of images and videos. One can distinguish two classes: 2D attention models, which concern images, and 3D models (2D +T) concerning videos. Although such a separation may seem artificial as the same principles govern the design of the models in both cases, we prefer to treat the spatio-temporal content separately.  
	
	\subsection{2D Attention Models}
	\label{subsection:2DAtt}
	
	One of the pioneering works introducing the use of an attention model in neural networks for image classification is presented in~\cite{Attention::squeeze}. The authors are interested in the contribution of feature channels along convolutional layers into decision making. The attention model here is ``global'': a channel weighting mechanism is introduced by ``attention blocks''. The processing consists of three steps: i) synthesis (\textit{squeeze}), ii) excitation (\textit{excitation}) and iii) feature scaling (\textit{scale}). Thus, for each channel, a block is a small network of neurons that learns a weighting coefficient. The next layers of the network ingest the characteristic channels thus weighted.
	This global weighting has been used as a basis for the authors of~\cite{Attention::A2Nets} who propose ``double attention'' blocks, i.e. ensuring a global and spatial weighting of the characteristics in convolution network layers.
	The authors of~\cite{Attention::RANImageClassification, Attention::Res3ATN} use the principles of residual neural networks to propose ``residual'' learning of the attention masks incorporated in the convolution layers. Their experiments on CIFAR data bases show that on CIFAR 10 the residual attention network with depth of $452$ has the best error rate compared to all the basic residual networks ($3.90$\%). The authors propose the incorporation of attention mechanisms in both forward (\textit{forward}) and backward (\textit{backward}) runs. This is also the approach we had in~\cite{Attention::saliency}, but by selecting important characteristics and not by weighting features and channels. Note that when minimizing the objective function by gradient descent, the attention mechanisms are implicitly introduced via the derivative calculation where the weighted characteristics are used. The authors of~\cite{Attention::RANImageClassification} report that this use in back propagation makes the training data robust to noise. This is also our approach in this work.
	\par Other works such as~\cite{Attention::Transfer} propose ``Teacher-student'' networks where the ``Teacher'' network is the one that learns attention and guides the student network for the image classification task. In our approach we also use a kind of attention transfer as in our architecture the attention branch and trunck brunch will join together for selection of important features. Also, in~\cite{Attention::EmbeddingImageWord}, attention mechanism is coupled with LSTM to learn the correlation between different data modalities such as text and image and therefore leads to better embedding.
	
	\subsection{3D Attention Models}
	\label{subsection:3DAtt}
	
	We focus here on the contribution of ``3D'' (2D+T) spatio-temporal attention models in deep networks for the action recognition problem.
	\par
	Attention mechanism is used in~\cite{Attention::skeleton} on joint skeleton and coupled with LSTM for 3D action recognition task. They reported better accuracy with attention mechanism than without. They also propose a recurrent attention mechanism on their model which strengthens the attention effect but might not lead to better performances if iterated too many times. In~\cite{Attention::channelsTemporal}, the author consider attention on the channels of aggregated temporal features extracted from videos on appearance and motion streams. Similarly, the authors in~\cite{Attention::Pyramid} construct a spatial attention model for each image by introducing feature pyramids. The temporal extension is obtained by a simple aggregation of the attention maps estimated for each of the K images of the pyramid extended to the spatio-temporal domain. Motion information is not taken into account. We differ from this approach by introducing attention blocks in our twin network at the level of the two branches: RGB and the optical flow.
	\par
	In~\cite{Attention::DanceFlow}, motion information, via the optical flow, acts as the attention map for locating actions in the video. 
	The authors introduce the ``motion condition'' layer to train the network on RGB appearance components conditional to this optical flow based map. The motion weighting layer allows to modify the spatial characteristics in the convolution layers.
	We find here the philosophy of using motion as an indicator of areas of interest~\cite{Attention::MotionObjectDetection}. Once more, the difference of our approach consists in introducing attention blocks in the two branches (RGB and optical flow) of the twin network.
	\par
In \cite{Attention:AttentionClusterOnFeatures}, attention clusters over the temporal dimension are used on image features extracted using Inception-ResNet-v2 \cite{NN:Inceptionv3} for RGB and Flow modalities. The Inception-ResNet-v2 models are pretrained on ImagetNet \cite{Dataset:ImageNetchallenge} and are fine-tuned for the optical flow model. A third branch processes the audio signal using VGG-16 \cite{NN:VGG} on extracted spectrogram samples and is processed similarly to an image \cite{Features:AudioFeaturesMethod}.
	\par
	3D attention blocks have also recently been introduced in ResNet 3D type networks for the recognition of 3D hand gestures from videos~\cite{Attention::Res3ATN} or from action recognition dataset \cite{Attention::3DRANs} such as HMDB-51~\cite{Dataset:HMDB}, UCF-101~\cite{Dataset:UCF101} and Kinetics~\cite{Dataset:Kinetics}. The authors of \cite{Attention::Res3ATN} build on the work of~\cite{Attention::RANImageClassification}, and propose a convolution network using the RGB image for feature extraction, and another coupled network to determine a soft attention mask with the derivable sigmoid function. The values of the extracted mask are then combined with the characteristics extracted from the RGB array. As in~\cite{Attention::Pyramid}, the authors do not use the motion information explicitly.
	\par
	Hence, our approach differs from current methods in the literature in the following:
	\begin{itemize}
		\item we introduce the 3D attention blocks into the two video streams: the branch containing the spatial information (RGB) and the branch containing the temporal information (optical flow).
		\item movement (optical flow) plays the discriminating role in our fine-grained classification context, our problem being to recognize actions and not to locate them.
	\end{itemize}
	In the following section, we first introduce the basis of our approach - the 3D (2D+T) convolutional twin arrays that we previously proposed in~\cite{PeCBMI}, we then detail the spatio-temporal attention blocks introduced in our network.
	
	\section{3D Attention Mechanism in Twin Space-Time Networks}
	\label{section:method}
	
	We first introduce our Twin Spatio-Temporal Convolutional Neural Network - TSTCNN used for classification and then detail the 3D attention and residual blocks developed and tested. 
	
	\subsection{Twin Spatio-Temporal Convolutional Neural Network - TSTCNN}
	\label{section:architecture}
	
	\begin{figure*}[htbp]
	    \vspace{2mm}
		\centering
		\includegraphics[width=\linewidth]{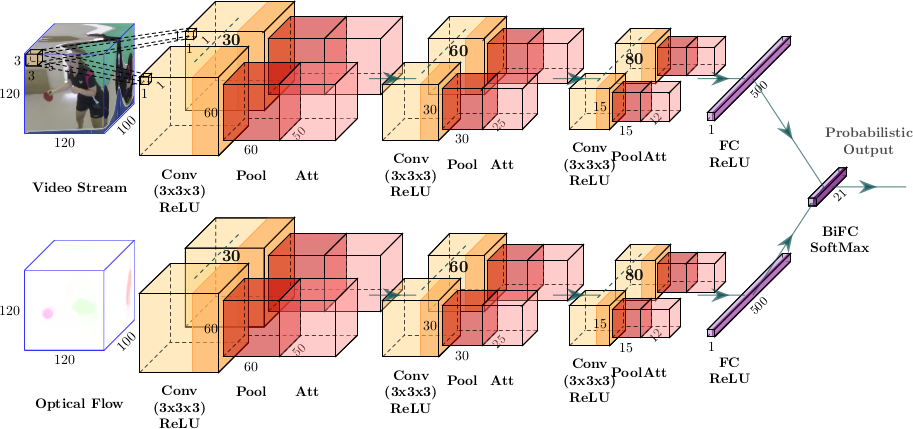}
		\caption{Twin network architecture with spatio-temporal convolutions and attention mechanism. The number of filters for each convolution are depicted above them.}
		\label{fig:twin}
	\end{figure*}
	
	In order to perform action classification in videos, we use a two streams convolutional neural network (\textit{twin}) with attention mechanism. Its architecture without attention blocks and early results on \texttt{TTStroke21} are described in~\cite{PeMTAP} and the analysis of the impact on classification of optical flow normalization is done in~\cite{PeICIP}. Its architecture with attention mechanism is presented in figure~\ref{fig:twin}. The difference from other Two stream networks~\cite{NN:SimonyanTwoStream, NN:TwoStreamFusion, NN:3dObjectPoposal}
	lies in: i) the symmetries of our network, ii) the input 4D data type (horizontal, vertical, temporal and channel) and iii) the final fusion step with a bi-linear layer at the end of our two branches.
	\par
	Our twin network, the so-called \texttt{TSTCNN} - (Twin Spatio-Temporal Convolutional Neural Network) consists of two individual branches: one branch takes as input the values of the RGB images of the sequence, the other branch uses the optical flow estimated by the method of~\cite{OF:BP}. It thus allows the incorporation of both spatial and temporal features~\cite{PeCBMI}. The played stroke is predicted from the RGB images of the sequence and the estimated motion vectors $ {\bf v} = (v_x, v_y)^T $.
	\par
	Each branch consists of three convolutional layers comprising successively $30$, $60$ and $80$ 3D filters, followed by a fully connected layer of size $500$. The 3D convolutional layers use space-time filters of size $3 \times 3 \times 3$. The two branches are merged through a final bilinear fully connected layer of size $21$, followed by a Softmax function to obtain an output class membership probability.
	A detailed implementation and description of our network is available on GitHub \footnote{\url{https://github.com/P-eMartin/crisp}} to facilitate reproducibility or the use of our method for other applications.
	
	\subsubsection{Learning Phase}
	\label{subsection:Datal_Aug}
	
	learning of our \texttt{TSTCNN} network is done by stochastic gradient descent (SGD) with Nesterov momentum~\cite{Deep::nesterov}.
	In order to avoid overfitting, data augmentation is performed in the spatial domain using rotations, homotheties and scale transformations. Data augmentation is also performed in the time domain in order to add variability around the temporal boundaries of the played stroke. We refer the reader to~\cite{PeCBMI} for more details. 
	
	\subsubsection{Performance without attention} the classification on the test data of the database \texttt{TTStroke-21} gives an accuracy of $91.4$\% in~\cite{PeCBMI}. For comparison, the two-branch I3D model~\cite{NN:I3DCarreira}, used as a baseline, gives an accuracy of only $43.1$\% for the same dataset using a temporal window of $64$ frames. Even if the results are satisfactory, it should be noted that the trained network gives comparable results in terms of accuracy: the gain compared to a network using image information only is $2.7$\%. There are therefore still room for improvement, notably by making better use of motion information during the training phase, as in~\cite{NN:Laptev18, PeICIP}, or by introducing information from 3D attention models.
	\par
	Detailed results, including experimental settings and further analysis, are presented in~\cite{PeCBMI}.
	
	\subsubsection{TSTCNN with residual and attention mechanism}
	to test the efficiency of residual and attention blocks, we first added residual block after the max pooling layers starting from the first max pooling layer until reaching all the max pooling layers. We did the same with attention blocks so to see the impact of each type of blocks and the impact of their number in the network. We also experimented using separated branches of the network and training them separately, RGB branch denoted as ``RGB'' model and Flow branch denoted as ``Flow'' model, to analyse the contribution on each stream. The TSTCNN, denoted as ``Twin'' model, with 3 attention blocks, is presented on figure~\ref{fig:twin}.
	
	\subsection{3D attention block}
	\label{sec::attention}
	
	\begin{figure*}[t!]
	    \vspace{2mm}
		\centering
		\includegraphics[width=\linewidth]{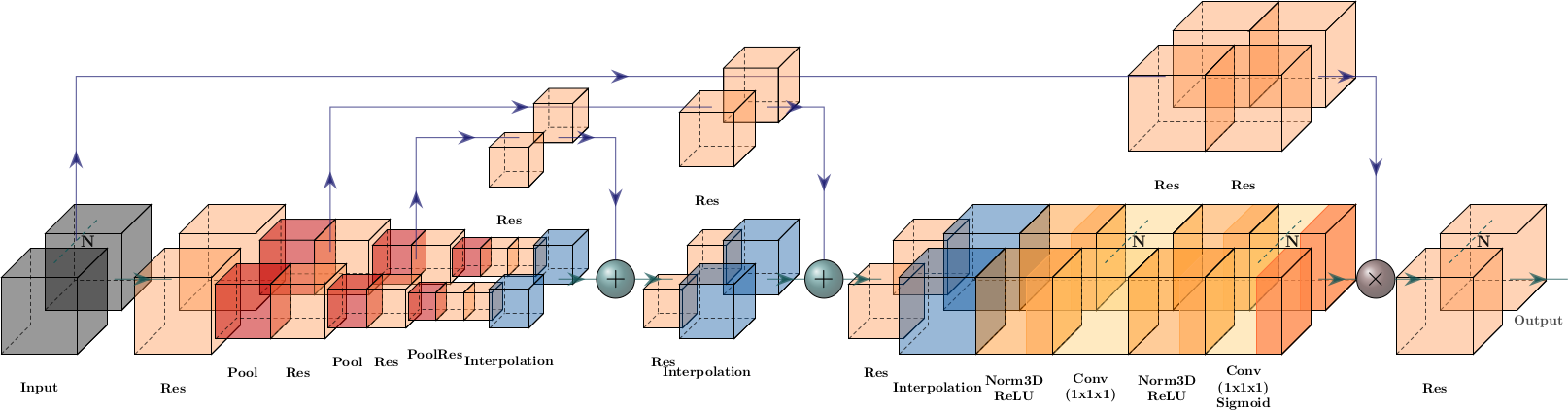}
		\caption{3D attention block architecture}
		\label{fig:attention}
	\end{figure*}
	
	3D attention block, inspired by the work carried out in 2D~\cite{Attention::RANImageClassification}, takes as input a 4D data block of size ($N \times W \times H \times T$) as illustrated in Figure~\ref{fig:attention}. In this block, all convolution presented uses the same number of filters, $N$, to maintain the dimension of the processed data. Our input data are processed by a first 3D residual block, denoted as ``\emph{res}'', presented in section~\ref{sec::resbloc}. Our network then splits in two branches: the trunk branch consisting of 2 successive 3D residual blocks (equation~\ref{eq:trunkbranch}) and the soft floating mask branch (lowest position in Figure~\ref{fig:attention}), described through the equations~\ref{eq:floatingbranch1}, \ref{eq:floatingbranch2}, \ref{eq:floatingbranch3}. Its role is to accentuate the features generated by the trunk branch. Those two branches are merged as described in equation~\ref{eq:attbranches}.
	
	\begin{equation}
    \label{eq:trunkbranch}
        branch_{trunk}(.) = res(res(.))
    \end{equation}
	
	The soft mask branch is constituted of several 3D residual blocks followed by Max Pooling layers, denoted as ``\emph{MaxP}''. It increases the reception field of convolutions using a bottom-up architecture, denoted as $f_{bu}(.)=res(MaxP(.))$. The lowest resolution is obtained after $3$ Max Pooling steps.
	
	\begin{equation}
    \label{eq:floatingbranch1}
    \begin{array}{l}
        x_1 = f_{bu}(res(Input))\\
        x_2 = f_{bu}(x_1))\\
        x_3 = f_{bu}(x_2))
    \end{array}
    \end{equation}
	
	The information is then extended by a symmetrical top-down architecture, $f_{td}(.) = Inter(res(.))$, to project the input features of each resolution level. ``\emph{Inter}'' denotes the trilinear interpolations~\cite{3DInterpolation} used for up-sampling. Two skipped connections are used for collecting information at different scales.
	
	\begin{equation}
    \label{eq:floatingbranch2}
    \begin{array}{l}
       y_1 = f_{td}(x_3) + res(x_2)\\
        y_2 = f_{td}(y_1) + res(x_1)\\
        y_3 = f_{td}(y_2)
    \end{array}
    \end{equation}

	The soft mask branch is then composed of 2 successive layers. Each includes a 3D batch normalization, denoted as $F_n(.)$ as described by equation~\ref{eq:batchnorm3d}, followed by a ReLU activation function and a convolution layer with kernel sizes ($1\times1\times1$). This is expressed by equation~\ref{eq:fconv}:
	
	\begin{equation}
    \label{eq:fconv}
        f_{conv}(.) = conv(ReLU(F_n(.)))
	\end{equation}
	
	It ends with a sigmoid function, denoted as ``Sig'', to scale values between $0$ and $1$. These two layers are depicted on the right of the lowest branch in figure~\ref{fig:attention} and are expressed by equation~\ref{eq:floatingbranch3}. 
	
	\begin{equation}
    \label{eq:floatingbranch3}
        branch_{fmask}(Input) = Sig(f_{conv}(f_{conv}(y3)))
    \end{equation}
	
	The output of our trunk branch is then multiplied term by term by $(1 \oplus  branch_{fmask} (Input))$ where $branch_{fmask}(Input)$ is the output of the mask branch. The result is then processed by the last 3D residual block $res(.)$ which ends the attention block, see equation~\ref{eq:attbranches}. 
	
	\begin{equation}
    \label{eq:attbranches}
        y = res(branch_{trunk}(Input) \odot (1 \oplus branch_{fmask}(Input)))
    \end{equation}
    Here the $\odot$ is an element-wise multiplication and $\oplus$ is an addition of a scalar to each vector component.
    
	\subsection{3D residual block}
	\label{sec::resbloc}

\begin{figure}[h]
		\centering
		\includegraphics[width=\linewidth]{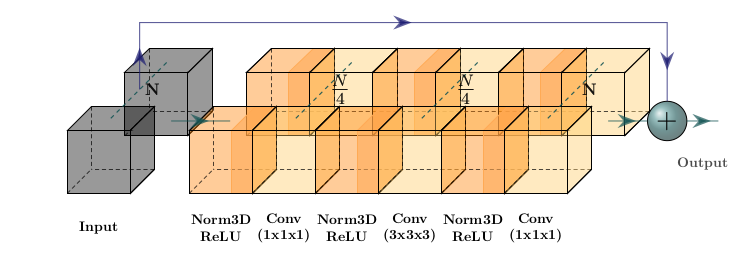}
		\caption{3D residual block architecture}
		\label{fig:resblock}
	\end{figure}

     Implemented 3D residual block (Fig.~\ref{fig:resblock}) inspired by the work carried out in 2D in~\cite{NN:ResNet}, takes as input a 4D data block of size ($N \times W \times H \times T$) representing respectively the number of channels, the two spatial dimensions and the temporal dimension. Input data are then processed by 3 successive layers $f_{conv_i} , i=1,...,3 $ (eq.~\ref{eq:floatingbranch3}). The result of residual block is the sum of the output of these $3$ successive layers and our input data:
	
	\begin{equation}
    \label{eq:res}
        res(x) = f_{conv_3}(f_{conv_2}(f_{conv_1}(x))) +  x
    \end{equation}
    
	Here, the first layer $f_{conv_1}$ uses $\frac{N}{4}$ convolution filters of size ($1\times1\times1$), the second layer $f_{conv_2}$ uses $\frac{N}{4}$ convolution filters of size ($3\times3\times3$). Finally, the third layer $f_{conv_3}$ employs $N$ convolution filters of size ($1\times1\times1$).
	
	The 3D batch normalization, described in~\cite{BatchNorm3D}, is performed channel by channel over the batch of data. If we have $x = (x_{1}, x_{2}, ... , x_{N_{channels}})$, then the normalization is $F_n(x) = (f_n(x_{1}), f_n(x_{2}), ... , f_n(x_{N_{channels}}))$ with:
	
	\begin{equation}
    \label{eq:batchnorm3d}
        f_n(x_i) = \frac{x_i - \mu_i}{\sqrt{\sigma_i^2 \oplus \epsilon}} * \gamma_i \oplus \beta_i
    \end{equation}
    
    with $i = 1, ... N_{channels}$, $\mu_i$ and $\sigma_i$ the mean and standard deviation vectors of $x_i$ computed over the batch, $\gamma_i$ and $\beta_i$ learnable parameters per channel and the division by $\sqrt{\sigma_i^2\oplus\epsilon}$ is element-wise. Here, $N_{channels}=N$ or $\frac{N}{4}$, depending on the normalization position in the residual block.
	
	\section{Experiments and results}
	\label{section:Results}
	
	To assess the efficiency of the attention block for capturing qualitative features for classification task, we compare the classification results of the model with and without attention blocks on the \texttt{TTStroke-21} dataset~\cite{PeCBMI}.
	
	\subsection{\texttt{TTStroke-21} dataset}
	
\begin{figure}[htb]
		\centering
		\begin{tabular}{cc}
			\includegraphics[height=.25\linewidth]{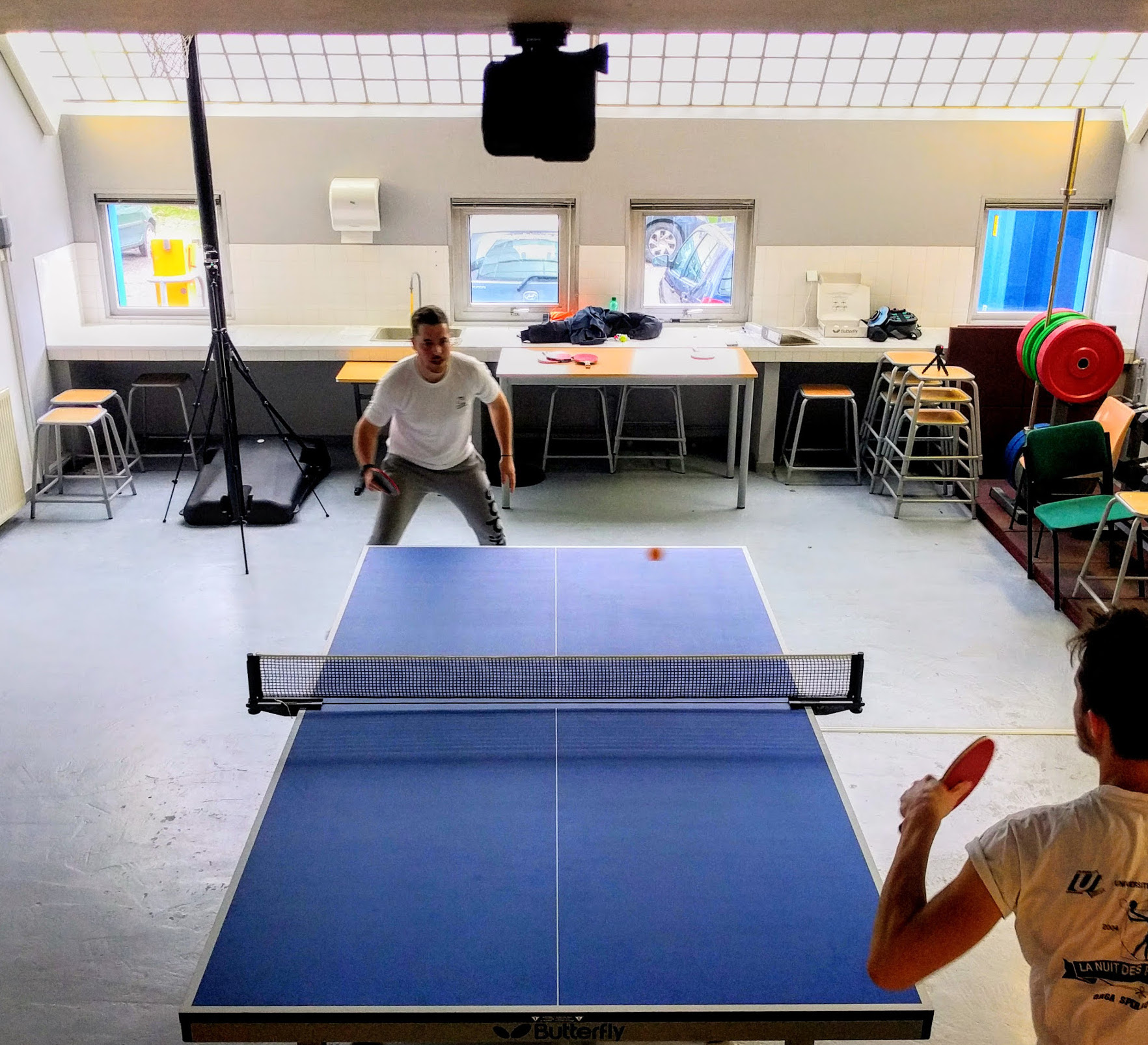}&
			\includegraphics[height=.25\linewidth]{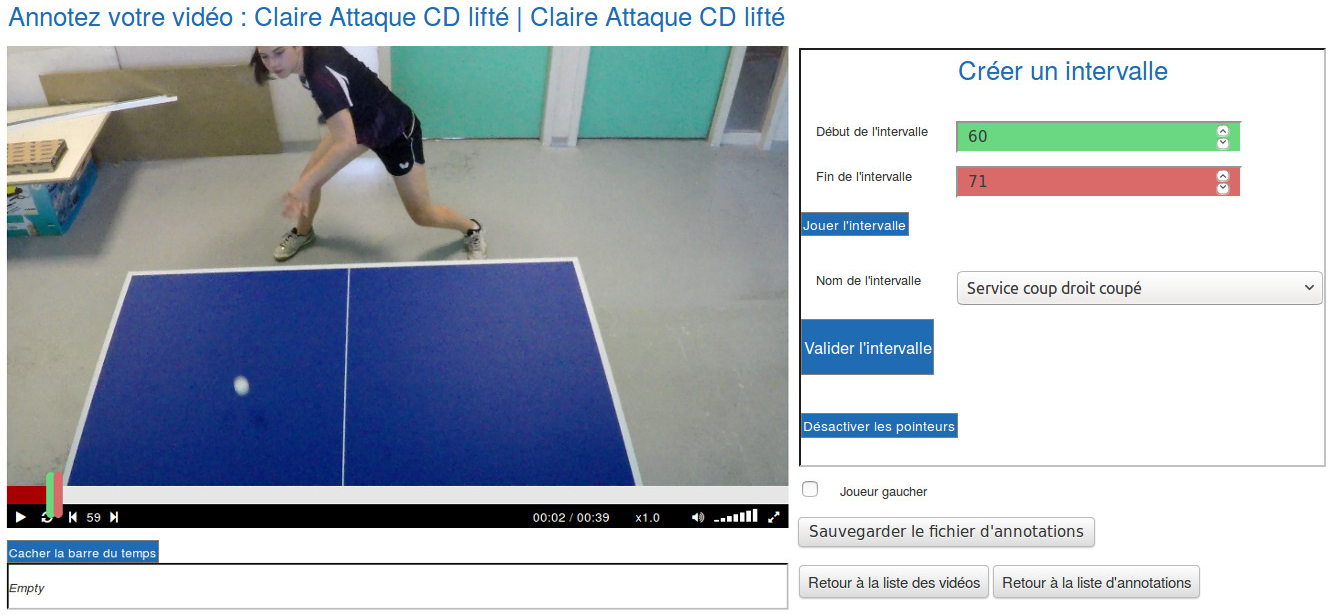}\\
			\textbf{a.} Video acquisition&
			\textbf{b.} Annotation platform\\[4pt]
		\end{tabular}
		\caption{\texttt{TTStroke-21} acquisition and annotation platform}
		\label{fig:dataset}
	\end{figure}

	Our dataset, entitled \texttt{TTStroke-21}~\cite{PeCBMI}, is composed of recorded table tennis game videos. These sequences are recorded indoors at different frame rates, with artificial light and without markers. The player is filmed in game or training situations, see figure~\ref{fig:dataset}a. These videos have been annotated by table tennis players and experts from the Faculty of Sports (STAPS) of the University of Bordeaux, France. A web platform  has been developed by our team for this purpose where the annotator locates in time and labels the strokes performed. The annotation platform is presented in figure~\ref{fig:dataset}b.
	
	\par
	In the following experiments, $129$ of videos recorded at $120$ frames per second have been considered. They represent a total of $1048$ table tennis strokes. From these time-segmented table tennis strokes in these $129$ videos, $681$ negative samples were selected but only $106$ were used to maintain class balance.
	
	\subsection{Visualizing the impact of the attention mechanism on features}
	
\begin{figure}[htb]
	\vspace{2mm}
		\centering
		\includegraphics[width=.9\linewidth]{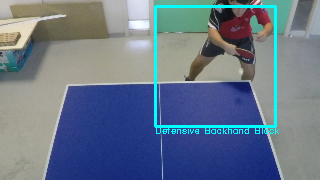}\\
		\textbf{a.} RGB \\[4pt]
		\begin{tabular}{ccc}
		   \includegraphics[width=.25\linewidth]{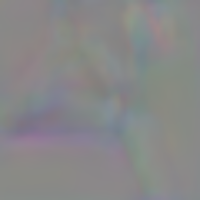} & \includegraphics[width=.25\linewidth]{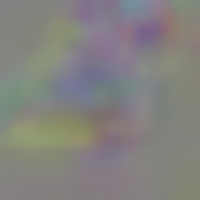} & \includegraphics[width=.25\linewidth]{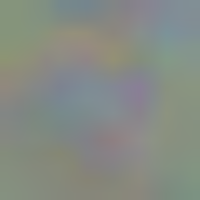}\\
		   \textbf{b.} Att1 & \textbf{c.} Att2 &  \textbf{d.} Att3\\
		\end{tabular}
		\caption{Visualisation of soft mask branch output from each attention block of the RGB branch of the ``Twin'' model. The RGB input segmented from the original frame and its class is represented in light blue on \textbf{a}.}
		\label{fig:attentionvisu}
	\end{figure}

	The attention block highlights features that contribute the most to the classification. In this way, the model can learn faster meaningful features in the  classification task. Figure~\ref{fig:attentionvisu} shows outputs of floating mask branch of each attention block for a RGB image input to the ``Twin'' model (Fig.~\ref{fig:twin}). Feature values range from $0$ to $1$, but are normalized using min-max normalization and resized for better visualization.
	
	RGB input is of size ($100\times120\times120$), each dimension representing respectively time, width and height. These parameters were fixed experimentally as a function of video resolution, frame-rate and stroke speed. The output size of the soft mask branch of attention blocks decreases by a factor $2$. Figure~\ref{fig:attentionvisu}b, c and d  represents 3 channels at a specific time. It can be noticed how the network focuses on the table's edges, on the player and even on the ball (more visible in figure~\ref{fig:attentionvisu}c). To classify a stroke, it is important to observe the posture of the player but also his/her position with respect to the table. The ball position and trajectory can also be of high importance to classify the stroke. On the whole training set, output values of the soft mask branch range between $0.35$ and $0.65$, meaning no features are totally left out or overrated, on the contrary.
	
	\subsection{Convergence of the models}
	\label{subsection:convergence}
	
	Conducted experiments required to change the hyper parameters used in our previous work~\cite{PeCBMI}. Indeed, the number of parameters to train, which depends on the number of attention or residual blocks, greatly increased compared to our first experiments without attention mechanisms, see table~\ref{tab:nb_parameters}.
	
	\begin{table}[htb]
	    \centering
	    \caption{Number of parameters to learn according to the architecture of the model$^*$ }
	    \begin{tabular}{c|c|c}
	        Models & without attention blocks & with 3 attention blocks \\ \hline
	        RGB & $180$ $800$ & $498$ $420$ \\[1pt]
	        Flow & $179$ $990$ & $497$ $610$\\[1pt]
	         Twin &  $360$ $790$ & $996$ $030$ \\[1pt]
	    \end{tabular}
	    \\[3pt]
	    $^*$ parameters of the fully connected layers are not considered.
	    \label{tab:nb_parameters}
	\end{table}
	\par
	We compared our results with the I3D model~\cite{NN:I3DCarreira} which contains around 25M parameters to train. Their model uses Inception modules introduced in~\cite{NN:GoogLeNet} which are combination of different 3D convolutional layers using different filter sizes and concatenating their output. We trained their model according to their instructions with rgb data and optical flow trained separately. The training process differs according to the type of data: a larger number of iterations is required for optical flow, with a specific scheduled learning rate. The output of the two models on the test set can be combined together to improve the performances as shown in table~\ref{table:acc_all}. We train their model using a time window of $T=100$ frames which we have selected after several experiments. Note that in our previous work, $T$ was set to $64$ frames and therefore performances were limited.
	\par
	Furthermore, in our case the type of model trained (``RGB'', ``Flow'' or ``Twin'') also influences the training process. Since different combinations and number of blocks were tested, the learning rate during training had to be adapted. A learning rate scheduler was then used, which reduced and increased the learning rate when the observed metric reached a plateau. Weights and state of the model were saved when it was performing the best and we re-loaded when the learning rate was changed. This allowed to re-start the optimization process from the past state with a new step-size in the gradient descent optimizer.
	\par
	We started training with a learning rate of $0.01$.	A number of epochs: $patience$, set to $50$, was considered before updating the learning rate, unless the performance drastically dropped (in our case: $0.7$ of the best validation accuracy obtained).
	\par
	The metric of interest was the training loss: if its average on the last $25$ epochs was greater than its average on the $35$ epochs before, the process was re-started from the past state and the learning rate divided by $10$ until reaching $10^{-5}$. After this step, the learning rate was set back to $0.01$ and process continued. These numbers of epochs were set empirically after preliminary experiments.
	
	\begin{figure}[htb]
		\centering
		\includegraphics[width=\linewidth]{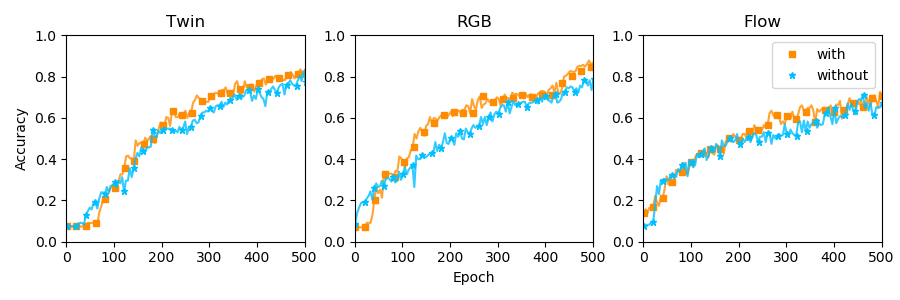}
		\caption{Evolution of the validation accuracy for the different models with and without attention blocks}
		\label{fig:performances}
	\end{figure}
	\par
	It worth mentioning that convergence is slower when using this method of leaning rate re-scheduler with our past architectures introduced in~\cite{PeCBMI, PeMTAP}. Here, this strategy is however efficient to adapt and find adequate learning rates during training for different architecture configurations.
	\par
	When comparing models with and without attention blocks as illustrated in~\ref{fig:performances}, it can be noticed, that our training process requires less epochs to adapt to our models with attention blocks. The convergence is faster and after the same number of epochs ($500$) models with attention blocks outperform models without attention (see table~\ref{table:acc_500}, where best results are depicted in bold for each type of model). 

    \begin{table}[htb]
		\centering
		\caption{Comparison of the classification performances after 500 epochs in terms of accuracy in \%}
		\label{table:acc_500}
		\begin{tabular}{c|cc|cc|cc}
            &\multicolumn{2}{c|}{I3D~\cite{NN:I3DCarreira}} &\multicolumn{2}{c|}{without attention} &\multicolumn{2}{c}{with attention} \\
			\hline
			Models & Train & Val. & Train & Val. & Train & Val. \\
			\hline
			RGB     &$95.1$   & $59.6$ &$83.9$	&$78.7$ &$96.3$	&$\bf87.8$	\\
			Flow    &$97.7$   & $58.7$ &$93$	&$71.8$ &$87.4$	&$\bf72.6$	\\
			Twin	&-  &-   &$88.9$	&$82.6$ &$92.7$	&$\bf83.5$ \\
		\end{tabular}
	\end{table}
	\par
	 Analysing results further, table~\ref{table:acc_500}, attention mechanisms seem to be more efficient with the one branch model ``RGB'' which is $11$\% better in term of classification accuracy.
	  With other models, the superiority is less obvious but still noticeable with $1$\% gain. We can notice the gap between training and validation accuracy for the I3D models. This model is far more deeper and has much more parameters to learn. It is  therefore more subject to over-fitting on our dataset. The attention blocks applied to our networks, even if  greatly increasing the number of parameters, seem to not over-fit. At $500$ epochs, the Twin model performs worse than the RGB model and still needs to be trained to outperform, see section~\ref{subsection:classificationperformances}. Its slower convergence can be due to the increased number of parameters to train. It may also come from the batch size used during training, which had to be decreased from $10$ to $5$ because of resource limitations.
	  
    \subsection{Classification performances}
    \label{subsection:classificationperformances}

    Our implemented attention blocks have shown to lead to faster convergence. The different models are also compared for the classification task. This comparison is done after reaching a stable convergence, which is around $1000$ epochs using the attention blocks and $1500$ without. All the results are reported in table~\ref{table:acc_all}), along with a comparison with the I3D model~\cite{NN:I3DCarreira} and the model introduced in our last work~\cite{PeICIP}. The Twin model without attention was retrained using the method presented in section~\ref{subsection:convergence} and better performances were also noticeable.
	 
	\begin{table}[htb]
		\centering
		\caption{Comparison of the classification performances after convergence in terms of accuracy \%}
		\label{table:acc_all}
		\begin{tabular}{cc|cc}
		Models & Train & Val. & Test \\
 		\hline
        RGB - I3D~\cite{NN:I3DCarreira}  &$98$   &$72.6$ & $69.8$\\[1pt]
        RGB~\cite{PeICIP} 		&$98.6$	&$87$	&$76.7$\\[1pt]
		RGB with Attention	&$96.9$	&$88.3$	&$\bf85.6$\\[3pt] 
		
		Flow - I3D~\cite{NN:I3DCarreira} &$98.9$ &$73.5$ & $73.3$\\[1pt]
		Flow~\cite{PeICIP}  	&$88.5$	&$73.5$	&$74.1$\\[1pt]
		Flow with Attention 	&$96.4$	&$83.5$	&$\bf79.7$\\[3pt] 
		
		RGB + Flow - I3D~\cite{NN:I3DCarreira}   &-   &-  & $75.9$\\[1pt]
		Twin~\cite{PeICIP}  &$89.4$	&$79.1$	&$72.4$\\[1pt] 
		Twin (Retrained)  &$99$	&$86.1$	&$81.9$\\[1pt] 
		Twin with Attention  &$97.3$	&$87.8$	&$\bf87.3$\\ 	
	\end{tabular}
	\end{table}
	
	\par
	The I3D model still suffers from over-fitting but its performances have greatly improved compared to our previous work~\cite{PeMTAP}. The reason is that the temporal dimension of input data increased from $64$ frames to $100$ frames. The combination of RGB and Flow models allows a $3$\% rise of the accuracy on the test set compared to the RGB model alone. The use of attention blocks allows to gain up to $12$\% with our RGB model and $5$\% with our Flow and Twin models. It may be surprising that the Twin could not benefit more from the attention mechanism contrary to its contribution for the other models. From the visualization analysis, figure~\ref{fig:attentionvisu}, and results in table~\ref{table:acc_all}, it can be argued that since the attention mechanism learns to focus on areas where RGB data are changing with respect to time, its contribution is lesser for models fed with  temporal information such as the optical flow.

	\section{Conclusion}
	\label{section:conclusion}
	
    In this work, we have extended the work carried out in 2D~\cite{NN:ResNet, Attention::RANImageClassification} to implement 3D residual blocks and 3D attention blocks. We applied these architectures to fine grained action recognition in video on \texttt{TTStroke-21}.
    \par
    We have shown that 3D attention blocks enable faster convergence of the models in terms of epochs. They outperform the models without attention blocks compared to our previous work with twin deep neural networks. However the amount of parameters to learn increases and the size of the network increases too, slowing down the training process for each epoch.
    The new training method also improves our Twin model without attention and performances of our baseline increased when considering longer input samples. After convergence, the Twin model with attention mechanism outperforms all other modalities.
    According to the visualization of the soft mask branch, it is safe to say that the attention blocks focus on meaningful features such as motion, body parts, position of the player with respect to the table, rackets and ball. We also noticed a greater efficiency of the attention mechanism on RGB data.
    \par
    We are limited by the number of samples of our datasets and we are still working on enriching \texttt{TTStroke-21} to add shades to our results. To deeper analyse the contribution of the attention blocks, we provide our implementation online and we are planning to test it on other publicly available datasets for fine grained action recognition~\cite{Dataset:Gym}.
	
	\section*{Acknowledgment}
	The work has been supported by research Grants CRISP of New Aquitania Region.

	\bibliographystyle{IEEEtran}
	\bibliography{biblio}
	
	
\end{document}